\documentclass[10pt,twocolumn]{article}

\usepackage[margin=0.78in]{geometry}
\usepackage{graphicx}
\usepackage{amsmath}
\usepackage{amssymb}
\usepackage{booktabs}
\usepackage{caption}
\usepackage{url}
\usepackage{microtype}
\usepackage[hidelinks]{hyperref}

\title{\bfseries Synthetic Leprosy Image Generation Using \\
Mask-Conditioned Latent Diffusion and Transfer Learning \\
from Large Chronic Wound Datasets}

\author{Yusuf Abdulkadir\\[2pt]
\normalsize Poolesville High School, Poolesville, Maryland, USA\\
\normalsize \texttt{yusufaae09@gmail.com}}

\date{}

\begin{document}

\twocolumn[{%
\begin{@twocolumnfalse}
\maketitle
\begin{abstract}
\noindent
Machine learning for neglected tropical diseases is limited by data, not
algorithms: public annotated image sets for leprosy (Hansen's disease) number in
the hundreds. We ask whether a model trained on abundant chronic wound photography
transfers to this low-data regime. We build a three-stage pipeline: a
DeepLabV3-ResNet50 segmenter supplies lesion masks for two wound datasets that ship
without them; a mask-conditioned latent diffusion model built from Stable Diffusion
1.5 components is trained on $3{,}280$ wound crops; and the model is fine-tuned on
leprosy image--mask pairs from CO2Wounds-V2 ($764$ images of $96$ patients). We
evaluate on $136$ held-out leprosy lesions using LPIPS anchored by a
real-versus-real baseline, kernel inception distance (KID), and a paired design
that runs both models on the same input. Fine-tuning moved the generated
distribution toward real held-out leprosy images: KID fell from $0.167$ to $0.089$,
and LPIPS distance to real images from $0.806$ to $0.765$, with non-overlapping
95\% confidence intervals. Giving the base model the leprosy prompt alone did not
reproduce the gain (KID $0.250$). For a fixed input, fine-tuning changes the output
less than a new random seed does (paired LPIPS $0.544$ versus $0.728$): it keeps
layout and alters tissue color and texture. The generated set shows no mode
collapse (internal diversity $0.734$ versus $0.667$ for real images). A dataset
audit shows that a majority of training images carry synthetic flat regions from
preprocessing, which the fine-tuned model partly reproduces. Chronic wound
photography is a viable donor domain for leprosy lesion synthesis; the remaining
barrier is semantic control rather than image quality.
\end{abstract}
\vspace{1.5em}
\end{@twocolumnfalse}
}]

\section{Introduction}

Leprosy is a chronic infectious disease caused by \textit{Mycobacterium leprae}
that attacks skin, peripheral nerves, eyes, and upper respiratory mucosa
\cite{fischer2017leprosy}. Roughly 200{,}000 new cases are reported each year,
concentrated in India, Brazil, and Indonesia. The disease is fully curable with
multidrug therapy, and the damage that makes it feared historically---deformity,
blindness---follows from undetected sensory loss rather than from the bacterium
directly. Early detection therefore matters enormously, and early detection is
substantially a visual task \cite{alrehaili2023leprosy}.

Machine learning ought to help. It largely has not, because the data is not
there. Leprosy is rare, and clinical images from endemic regions are seldom
collected and annotated for public release, so the datasets that exist are small.
The largest we are aware of, CO2Wounds-V2 \cite{sanchez2024co2wounds}, contains several hundred
images. Deep generative and diagnostic models are typically trained on orders of
magnitude more. This is the familiar predicament of the neglected tropical
diseases: the conditions with the least data are the ones where automated triage
would matter most.

One response is synthetic augmentation. If a generative model can learn what
lesions look like from a large, related, well-populated domain, and then be
adapted to a small target domain, the target's data ceiling becomes less binding.
Chronic wound photography is a natural donor domain. It is abundant, publicly
released, and shares the low-level statistics that matter---skin tone, granulation
and slough texture, irregular lesion boundaries, clinical lighting and framing.

This paper asks a narrow, testable version of that question:

\begin{quote}
\textit{Can a diffusion model pretrained on large chronic wound datasets and
fine-tuned on limited leprosy data generate synthetic leprosy images whose
perceptual distribution approaches that of real leprosy images?}
\end{quote}

Our contributions are:

\begin{enumerate}
\itemsep2pt
\item \textbf{A complete three-stage pipeline}---automatic mask generation,
mask-conditioned latent diffusion pretraining on $3{,}280$ wound crops, and
transfer to leprosy crops---described at a level of detail sufficient to
reproduce.
\item \textbf{A held-out evaluation anchored by a real-versus-real baseline.}
Without that reference, cross-set LPIPS values cannot be interpreted at all. With
it, plus KID and a paired same-input design, we show that fine-tuning moves the
generated distribution toward real held-out leprosy images while the generated set
keeps at least the real set's diversity.
\item \textbf{A preprocessing audit} showing that a majority of training images on
both sides of the transfer contain large synthetic flat regions, from letterbox
padding in the wound corpora and color-card redaction in the leprosy corpus, and
that the fine-tuned model reproduces some of them.
\end{enumerate}

This paper is the second iteration of this work. It retrains the model with a
held-out split and replaces the earlier iteration's evaluation with the held-out
protocol, KID, and paired comparisons; where the two differ, most notably on how
much fine-tuning changes the outputs, the results here supersede the earlier ones.

\section{Background and Related Work}

\subsection{Clinical background}

Leprosy presents along an immunological spectrum, conventionally the
Ridley--Jopling classification, running from tuberculoid (TT, strong cell-mediated
response, few sharply bounded hypopigmented patches, low bacterial load) through
three borderline forms (BT, BB, BL) to lepromatous (LL, weak response, numerous
symmetrical nodules and plaques, high bacterial load) \cite{fischer2017leprosy}.
Visual presentation therefore varies enormously across the spectrum, which is
precisely what makes both diagnosis and generative modeling difficult: a model
must cover several distinct morphologies from few examples.

A point of scope worth stating plainly at the outset. The leprosy corpus used
here, CO2Wounds-V2 \cite{sanchez2024co2wounds}, consists of \emph{chronic wounds
and ulcers photographed in leprosy patients}. These are largely neuropathic
ulcers arising from the sensory loss leprosy causes, not the hypopigmented
macules and plaques of early diagnosis. Our model therefore learns leprosy-related
chronic ulceration, which is a real and clinically meaningful presentation, but
not the full Ridley--Jopling spectrum. We return to this in
Section~\ref{sec:limitations}.

\subsection{Synthetic medical image generation}

Generative augmentation for skin imaging has a short but active history. DermGAN
\cite{ghorbani2019dermgan} synthesized clinical skin images with specified
pathology and location. WG2AN \cite{sarp2021wg2an} generated wound images with a
GAN and is the direct ancestor of the naming used here. More recently, diffusion
models have displaced GANs for this task: Basiri et al.\ \cite{basiri2023dfu}
synthesized diabetic foot ulcer images with a diffusion model, and Hamrani et al.\
\cite{hamrani2025beyond} applied self-attention diffusion to zero-shot ulcer
segmentation. Liu et al.\ \cite{liu2023chronic} pursued the complementary route of
semi-supervised augmentation for chronic wound assessment. On the segmentation
side, Wang et al.\ \cite{wang2020wound} and Anisuzzaman et al.\
\cite{anisuzzaman2022mobile} established that convolutional networks segment
wounds reliably enough to be deployed, which is the premise Stage 1 relies upon.

We are not aware of prior work applying diffusion-based transfer learning to
leprosy imagery specifically.

\subsection{Latent diffusion and spatial conditioning}

Denoising diffusion probabilistic models \cite{ho2020ddpm} learn to invert a fixed
Gaussian corruption process. Latent diffusion \cite{rombach2022ldm} performs this
in the compressed latent space of a pretrained autoencoder, cutting the cost of
high-resolution synthesis by roughly an order of magnitude; Stable Diffusion 1.5 is
the widely used instantiation and the backbone here. Text conditioning enters
through CLIP embeddings \cite{radford2021clip} and is strengthened at sampling time
by classifier-free guidance \cite{ho2022cfg}.

Spatial conditioning---forcing generated content to respect a given geometry---is
the harder problem, and the choice of mechanism turns out to matter a great deal
for our results. The cheapest option is to concatenate the conditioning map to the
noisy latent at the network input, widening the first convolution. The stronger
option is ControlNet \cite{zhang2023controlnet}, which injects conditioning
features into the UNet at multiple resolutions through zero-initialized
convolutions. ControlNet exists because input concatenation is known to be weak:
a signal introduced only at the input can be attenuated as it propagates. We
implemented a lightweight multi-scale encoder in this spirit but, as
Section~\ref{sec:residual} documents, the reported model was in fact trained with
concatenation alone, and its behavior is consistent with that.

Two further training refinements are used. Min-SNR loss weighting
\cite{hang2023minsnr} rebalances the per-timestep objective, which otherwise
over-weights very noisy timesteps and yields washed-out texture. DPM-Solver++
\cite{lu2022dpmsolver} replaces ancestral sampling with a higher-order ODE solver,
giving sharp samples in 20--30 steps rather than hundreds. Alternative
formulations of the denoising process itself, such as residual denoising diffusion
\cite{liu2024residual}, decompose generation into separate residual and noise
pathways; we retain the standard formulation here.

\section{Data}

\subsection{Chronic wound corpora}

Three public collections form the donor domain.

\textbf{Kaggle Wound Segmentation Images} \cite{kagglewoundseg} is itself an
aggregation of three sources, distinguishable by filename prefix in the
distributed correspondence table: FUSeg foot-ulcer images (\texttt{fusc\_},
$n=1210$), the Medetec wound archive (\texttt{medetec\_}, $n=374$), and WSNET
(\texttt{wsnet\_}, $n=1176$). All $2{,}760$ images ship with ground-truth binary
lesion masks.

\textbf{AZH Wound Database} \cite{wang2020wound,anisuzzaman2022mobile} contributes
$930$ images labeled by wound etiology (venous, diabetic, pressure, surgical) but
\emph{without} segmentation masks.

\textbf{WoundsDB} \cite{chronicwounddb} contributes $79$ RGB photographs extracted
from a multimodal capture archive, also without masks.

All images were converted to RGB and resized to $512 \times 512$ with Lanczos
resampling.

\subsection{Leprosy corpus}

The target domain is CO2Wounds-V2 \cite{sanchez2024co2wounds}, $764$ chronic wound
photographs from $96$ leprosy patients; multiple images may depict
the same patient or the same lesion at different times. Lesion masks were
rasterized from the dataset's own COCO polygon annotations ($971$ polygons) rather
than predicted, so the leprosy masks are ground truth. Images and masks were
resized to $512 \times 512$.

\textbf{Patient-level dependence.} Because $764$ images derive from $96$
patients, images are not independent. The corpus does not identify patients, so
the held-out split used for evaluation (Section~\ref{sec:runs}) is by image: the
same patient can appear in both the fine-tuning and held-out sets. We revisit this
in Section~\ref{sec:limitations}.

\subsection{Color calibration card removal}

Many CO2Wounds-V2 photographs include a physical color calibration card placed
beside the lesion for standardization. Left in place, such a card is a strong,
highly regular visual feature that a generative model will happily reproduce. We
removed it with a vision-language agent: a GPT vision-language model, accessed
through the OpenAI API, was prompted to return a bounding
box for the card (explicitly instructed not to confuse organic wound tissue with
the artificial card), and the box was then filled with the median of the
surrounding pixels sampled from four 20-pixel bands.

\begin{figure}[t]
\centering
\includegraphics[width=\columnwidth]{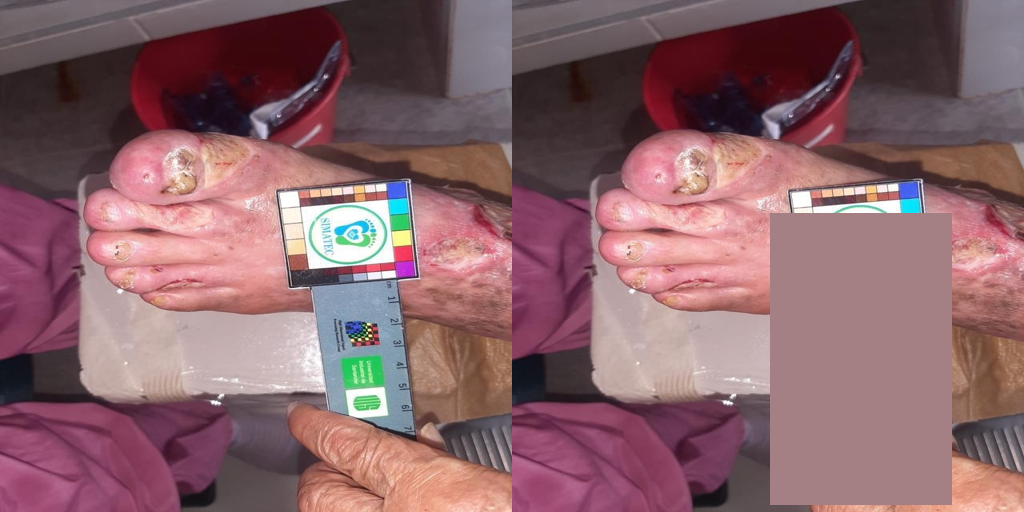}
\caption{Color calibration card removal (left: original, right: processed). The
card is replaced by a flat rectangle filled with the median surrounding color.
Two limitations are visible: the top strip of the card is not fully covered, and
the replacement region is perfectly uniform. This procedure affected $41.0\%$ of
the leprosy corpus.}
\label{fig:card}
\end{figure}

The procedure altered $313$ of $764$ images ($41.0\%$). It is imperfect in two
ways, both of which bear on the results. Detection
is sometimes incomplete---Figure~\ref{fig:card} shows a case where the upper strip
of the card survives the fill---and the fill itself replaces the card with a large
perfectly uniform rectangle, an artificial structure absent from any real
photograph.

\subsection{Dataset audit}
\label{sec:audit}

Before training we audited the assembled data. Two findings bear directly on the
results.

\begin{figure}[t]
\centering
\includegraphics[width=\columnwidth]{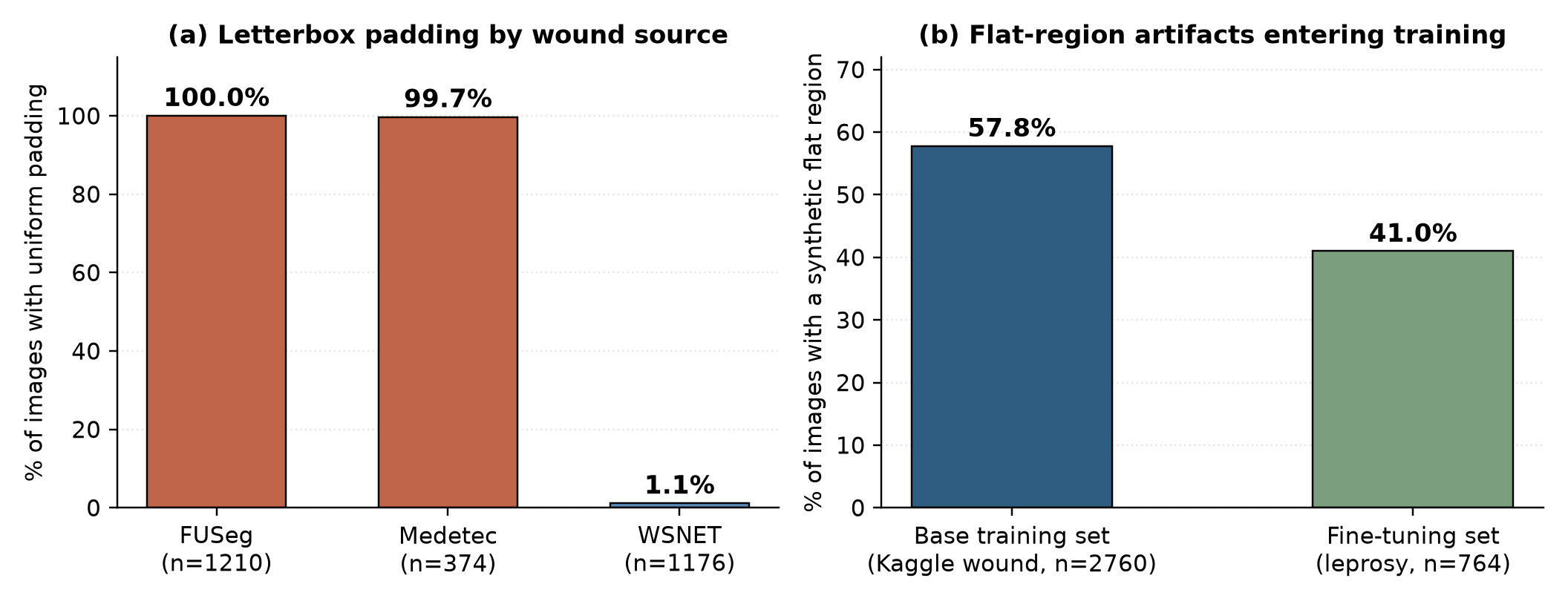}
\caption{Preprocessing artifacts in the training corpora. (a) Uniform letterbox
padding is near-universal in the FUSeg and Medetec sub-corpora and near-absent in
WSNET. (b) A majority of images on both sides of the transfer contain a large
synthetic flat region, from padding in the wound corpus and from card redaction in
the leprosy corpus.}
\label{fig:artifact}
\end{figure}

\textbf{Letterbox padding.} We flagged an image as padded when more than $20$ of its
$1024$ scanlines (rows plus columns) had per-channel standard deviation below $3$.
By this criterion, padding is essentially universal in two of the three Kaggle
sub-corpora and essentially absent from the third
(Figure~\ref{fig:artifact}a): $1210/1210$ FUSeg images ($100\%$), $373/374$ Medetec
images ($99.7\%$), and $13/1176$ WSNET images ($1.1\%$). Overall $1{,}596$ of
$2{,}760$ Kaggle images ($57.8\%$) carry uniform padding bars, covering on average
$13$--$15\%$ of scanlines. Combined with the $41.0\%$ card-redaction rate in the
leprosy corpus, a majority of images on both sides of the transfer contain large
synthetic flat regions (Figure~\ref{fig:artifact}b).

\begin{table*}[t]
\centering
\small
\caption{Dataset composition and attrition under the ROI filter. Mask source
indicates whether lesion masks are distributed with the corpus or predicted by our
Stage 1 network. \textit{Usable} counts the pairs that actually contributed
gradient during training.}
\label{tab:data}
\begin{tabular}{lrrrr}
\toprule
Corpus & Pairs & Empty & $<24$px & Usable \\
\midrule
\multicolumn{5}{l}{\textit{Base training (wound)}}\\
Kaggle (ground truth)  & 2760 & 24  & 217 & 2519 \\
AZH (predicted)        & 930  & 204 & 24  & 702 \\
WoundsDB (predicted)   & 79   & 5   & 15  & 59  \\
\textbf{Total}         & \textbf{3769} & \textbf{233} & \textbf{256} & \textbf{3280} \\
\midrule
\multicolumn{5}{l}{\textit{Fine-tuning (leprosy)}}\\
CO2Wounds-V2 (ground truth) & 764 & 27 & 29 & 708 \\
\bottomrule
\end{tabular}
\end{table*}

\begin{figure}[t]
\centering
\includegraphics[width=\columnwidth]{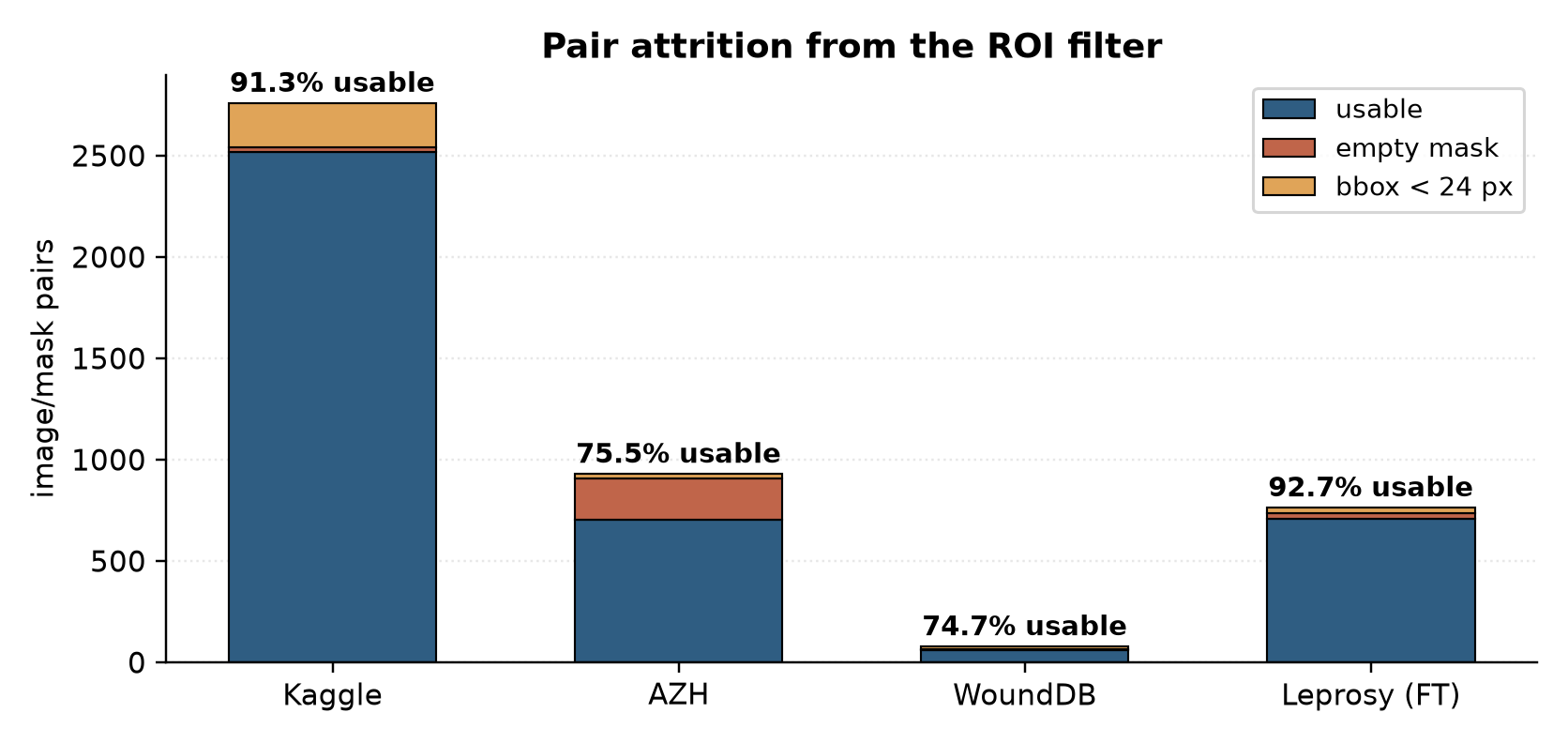}
\caption{Pairs discarded by the ROI filter. AZH loses $21.9\%$ of its pairs to
completely empty predicted masks, the clearest available signal of pseudo-label
degradation under domain shift.}
\label{fig:attrition}
\end{figure}

\textbf{Pair attrition.} The ROI filter described in Section~\ref{sec:roi} silently
discards pairs whose mask is empty or whose lesion bounding box has a side shorter
than $24$ pixels. Quantifying this (Table~\ref{tab:data},
Figure~\ref{fig:attrition}) shows the nominal and effective dataset sizes differ
substantially: of $3{,}769$ collected wound pairs, $3{,}280$ ($87.0\%$) contribute
gradient. The attrition is highly uneven. AZH is the worst case: $204$ of $930$
pseudo-masks ($21.9\%$) are entirely empty, meaning the segmentation network found
no wound at all. That figure is the most direct measure we have of pseudo-label
quality under domain shift, and it is a caution against assuming that a segmenter
validated on one corpus transfers silently to another.

\begin{figure*}[t]
\centering
\includegraphics[width=\textwidth]{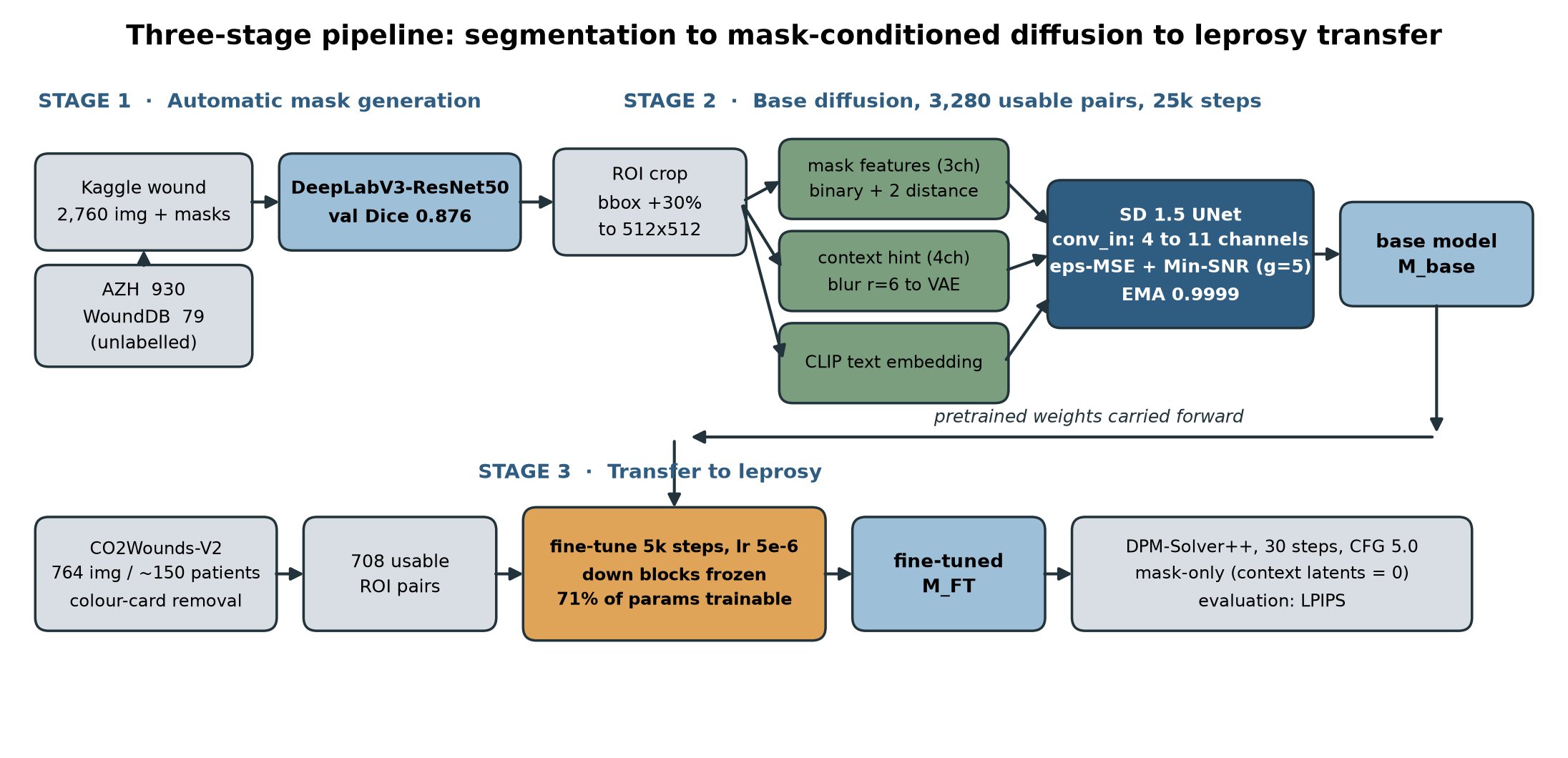}
\caption{The three-stage pipeline. Stage 1 trains a segmentation network on the
one wound corpus that ships with ground-truth masks and uses it to label the two
that do not. Stage 2 pretrains a mask-conditioned latent diffusion model on all
wound crops. Stage 3 transfers that model to leprosy. At sampling time the model
receives a lesion mask and Gaussian noise only; no real image pixels enter the
generation path.}
\label{fig:pipeline}
\end{figure*}

\section{Methods}

Figure~\ref{fig:pipeline} summarizes the pipeline. All work used PyTorch and the
HuggingFace \texttt{diffusers} library on AWS GPU instances (SageMaker for the first
training run and EC2 with an NVIDIA A10G for the second; Section~\ref{sec:runs}).

\subsection{Stage 1: automatic mask generation}

Two of the three wound corpora ship without masks, and the diffusion model requires
one per image. We therefore trained a segmentation network on the corpus that does
have ground truth and used it to label the others.

We fine-tuned \texttt{deeplabv3\_resnet50} \cite{chen2017deeplabv3,he2016resnet}
from COCO-pretrained weights, replacing the final classifier convolution with a
single-channel head and retaining the auxiliary head. Training used the $2{,}208$
image--mask pairs in the Kaggle training split at $512 \times 512$, with a random
$15\%$ validation partition (seed $42$). The objective combined binary
cross-entropy and soft Dice in equal parts,
\begin{equation}
\mathcal{L} = 0.5\,\mathcal{L}_{\text{BCE}} + 0.5\,(1 - \text{Dice}),
\end{equation}
with Dice smoothing $1.0$. A conservative augmentation was implemented---horizontal
flip ($p{=}0.5$), vertical flip ($p{=}0.2$), and mild color jitter (brightness and
contrast $0.15$, saturation $0.05$, hue $0.02$) on the image only---but because the
training and validation partitions shared one dataset object, switching it off for
validation also switched it off for training, so the reported network was trained
without augmentation. We optimized with AdamW \cite{loshchilov2019adamw} at learning rate
$10^{-4}$, weight decay $10^{-4}$, batch size $4$, mixed precision, for $40$ epochs
on an NVIDIA Tesla T4 (approximately $463$\,s per epoch, $5.1$\,h total). The
checkpoint with the best validation Dice was retained.

The trained network was applied at a probability threshold of $0.5$ to all $930$
AZH and $79$ WoundsDB images, producing binary masks for $1{,}009$ images
($26.8\%$ of the base training pairs). Kaggle and leprosy masks remained ground
truth throughout.

\subsection{Stage 2: mask-conditioned latent diffusion}

The generative model, which we call WG2Diff-ROI in acknowledgment of WG2AN
\cite{sarp2021wg2an}, is a latent diffusion model \cite{rombach2022ldm} built from
Stable Diffusion 1.5 components. The variational autoencoder (latent scaling
$0.18215$) and the CLIP ViT-L/14 text encoder are frozen; only the UNet is trained.

\subsubsection{Region-of-interest extraction}
\label{sec:roi}

Generating an entire clinical photograph wastes capacity on background. Instead we
train and sample on lesion-centered crops. For each pair we compute the mask
bounding box, expand it by $30\%$ of its width and height to retain surrounding
skin context, crop image and mask, and resize to $512 \times 512$ (bicubic for the
image, nearest-neighbor for the mask). Pairs with an empty mask or a bounding-box
side below $24$ pixels are skipped, with the consequences quantified in
Section~\ref{sec:audit}. Because the lesion now fills the frame, far more of the
$64 \times 64$ latent grid is spent on lesion detail.

\subsubsection{Conditioning signals}

Three signals condition the UNet.

\textbf{Mask features (3 channels).} Rather than a bare binary mask we supply the
binary mask together with two Euclidean distance transforms, one measuring distance
inward from the boundary and one outward, each normalized to $[0,1]$ per crop. The
distance maps give a smooth, signed sense of proximity to the lesion edge instead
of a discontinuous step, which sharpens generated boundaries.

\textbf{Context hint (4 channels).} The ROI crop is Gaussian-blurred with radius
$6$ and encoded by the frozen VAE. This carries low-frequency information---skin
tone, illumination direction, overall exposure---while destroying the lesion
texture the model must synthesize. To prevent dependence on a signal unavailable at
sampling time, the context latent is zeroed with probability $0.30$ during training.

\textbf{Text (77 tokens).} A fixed prompt is CLIP-encoded and supplied via
cross-attention; it is replaced by the empty string with probability $0.10$ so that
classifier-free guidance \cite{ho2022cfg} is available at sampling time. The base
model used \textit{``a realistic clinical photograph of a skin lesion.''}

\subsubsection{Architecture}

Mask features are nearest-neighbor downsampled to $64 \times 64$ and concatenated
with the noisy latent and the context latent, so the UNet input has
$4 + 3 + 4 = 11$ channels rather than $4$. We replace \texttt{conv\_in} with a
wider convolution, copying the pretrained weights into the first four input
channels and initializing the seven new channels from $\mathcal{N}(0, 10^{-3})$.
This preserves the pretrained prior at initialization while letting conditioning
influence grow during training.

\subsubsection{Objective and optimization}

Training follows the standard noise-prediction objective under a DDPM schedule
\cite{ho2020ddpm} with $1000$ timesteps, reweighted by Min-SNR
\cite{hang2023minsnr}:
\begin{equation}
\mathcal{L} = \mathbb{E}_{z_0, \epsilon, t}
\left[ w(t)\,\lVert \epsilon - \epsilon_\theta(z_t, t, c) \rVert_2^2 \right],
\end{equation}
\begin{equation}
w(t) = \frac{\min(\mathrm{SNR}(t), \gamma)}{\mathrm{SNR}(t)},
\qquad
\mathrm{SNR}(t) = \frac{\bar\alpha_t}{1 - \bar\alpha_t},
\end{equation}
with $\gamma = 5$.

We used AdamW at learning rate $10^{-5}$, weight decay $10^{-2}$, micro-batch size
$2$ with gradient accumulation $8$ (effective batch $16$), and fp16 mixed precision.
Images are sampled from the trained weights directly, without weight averaging.
Training ran for $25{,}000$ micro-steps, that is $3{,}125$ optimizer updates and
approximately $50{,}000$ image presentations, or $15.2$ epochs over the $3{,}280$
usable pairs.

\subsection{Stage 3: transfer to leprosy}

The fine-tuned model starts from the trained base weights in memory. All UNet
down-blocks are frozen on the reasoning that low-level wound texture transfers
directly and only the decoder path needs to adapt; this leaves
$609{,}928{,}644$ of $859{,}541{,}124$ UNet parameters trainable ($71.0\%$).
Learning rate is lowered to $5 \times 10^{-6}$, gradient accumulation reduced to
$4$, and the prompt changed to \textit{``a realistic clinical photograph of a
leprosy ulcer.''} Fine-tuning ran for $5{,}000$ micro-steps ($1{,}250$ optimizer
updates) on the same architecture and schedule.

\subsection{Training runs and held-out split}
\label{sec:runs}

The configuration above was trained twice. The first run fine-tuned on all $708$
usable leprosy pairs; its checkpoints were later lost, and its training curves,
sample images, and Kaggle-mask evaluation (Section~\ref{sec:kaggleprotocol}) are what
remain of it. The second run, which supplies all held-out results in this paper,
repeats every setting with one change to the data: the leprosy corpus is split by
image ($80/20$, seed $2027$) into $611$ fine-tuning images and $153$ held-out
images. After the ROI filter, $572$ pairs contribute to fine-tuning (about $17.5$
epochs) and $136$ held-out lesions are usable for evaluation. In the second run the
base and fine-tuned models are saved to separate locations, every generated image
records a hash of the weights that produced it, and the evaluation checks that the
fine-tuned weights started from the base weights and differ from them. The Stable
Diffusion 1.5 files were loaded from the
\texttt{stable-diffusion-v1-5/stable-diffusion-v1-5} mirror, since the original
repository is no longer hosted; its UNet parameter counts match the first run
exactly.

\subsection{Sampling}

At inference the model receives a lesion mask and nothing else drawn from a real
photograph. The mask is ROI-processed identically to training, the latent is
initialized from $\mathcal{N}(0, I)$, and the context latent is set to zero, which
the training-time context dropout has prepared the model for. Denoising uses
DPM-Solver++ multistep \cite{lu2022dpmsolver} for $30$ steps with classifier-free
guidance scale $5.0$, after which the frozen VAE decoder maps the final latent to a
$512 \times 512$ RGB image.

We emphasize the resulting claim precisely, since it is easy to overstate in either
direction. Real images are essential during \emph{training}. During
\emph{generation}, the only inputs are Gaussian noise, a binary geometry, and a
text embedding; no real image pixels enter the generation path. Outputs are
therefore synthesized rather than edited or retrieved. That the mask originated
from a real photograph does not change this, because a mask encodes shape and
location only---no texture, color, illumination, or patient-identifying content.

\subsection{Architecture progression}

The reported model is the third in a series, and the intermediate designs clarify
which choices mattered (Table~\ref{tab:designs}). Design A conditioned a
full-frame $512^2$ model on a single binary mask channel (5 input channels) and
produced soft, blurry lesions. Design B introduced ROI cropping and the
three-channel distance-transform mask representation (7 channels), which was the
single largest qualitative improvement: concentrating the latent budget on the
lesion sharpened texture markedly. Design C, reported here, added the blurred
context hint (11 channels) and widened the ROI margin from $20\%$ to $30\%$.

\textbf{Conditioning path in force.}\label{sec:residual} A lightweight
ControlNet-style encoder \cite{zhang2023controlnet} ($6{,}234{,}048$ parameters)
was implemented to inject mask features into the UNet at four resolutions ($64$,
$32$, $16$, $8$) through $1\times1$ projections matching the SD 1.5 block widths.
In the training and sampling routines executed for the reported runs, the UNet is
called without the \texttt{down\_block\_additional\_residuals} and
\texttt{mid\_block\_additional\_residual} arguments, so the encoder never entered
the computation graph and remained at initialization. The reported model
therefore conditions on lesion geometry through input concatenation alone---known
to be the weaker option, and the reason ControlNet exists. Design C's
contribution over design B is accordingly the context hint and the wider ROI
margin, not multi-scale conditioning, and Section~\ref{sec:discussion} traces the
model's loose mask adherence to this mechanism.

\begin{table}[t]
\centering
\small
\caption{Architecture progression. Only the final design was evaluated
quantitatively; earlier comparisons were qualitative.}
\label{tab:designs}
\begin{tabular}{llrr}
\toprule
Design & Framing & Mask ch. & UNet in \\
\midrule
A  & full frame       & 1 (binary)      & 5  \\
B  & ROI, $20\%$ margin & 3 (+ distance)  & 7  \\
C  & ROI, $30\%$ margin & 3 (+ distance)  & 11 \\
\bottomrule
\end{tabular}
\end{table}

\section{Preliminary Explorations}

The diffusion pipeline was not the first approach attempted. Two earlier
directions were pursued and set aside; both informed the final design, and we
summarize them because the reasons for abandoning them are part of the argument
for the method that succeeded.

\subsection{Class-conditional GAN synthesis}

The initial generative attempt was an auxiliary-classifier GAN
\cite{odena2017acgan} with a DCGAN backbone \cite{radford2016dcgan}, trained to
distinguish and generate two classes: leprosy images from CO2Wounds-V2 (label 0)
against a size-matched sample of other dermatological conditions drawn from
DermNet (label 1). The generator mapped a $100$-dimensional latent concatenated
with a learned class embedding through five transposed-convolution blocks to
$128 \times 128$ RGB; the discriminator carried separate adversarial and
classification heads. Training used Adam ($\text{lr}=2\times10^{-4}$,
$\beta_1 = 0.5$), batch size $32$, for $60$ epochs.

\begin{figure}[t]
\centering
\includegraphics[width=\columnwidth]{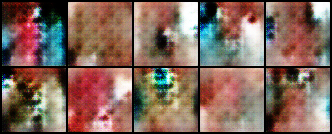}
\caption{Samples from the preliminary DCGAN at $128 \times 128$. Skin tone and
coarse lesion structure are present; fine texture and spatial control are not.
This limitation motivated the move to mask-conditioned latent diffusion.}
\label{fig:gan}
\end{figure}

A parallel unconditional DCGAN variant with spectral normalization in the
discriminator was also trained to $190$ epochs. Its samples
(Figure~\ref{fig:gan}) reproduce plausible skin tone and coarse lesion-like blobs
but lack the texture detail that distinguishes granulation from slough from
necrosis, and show the mode concentration typical of small-data GAN training.

Three limitations drove the switch to diffusion. Output resolution was capped at
$128 \times 128$ by training stability, well below what lesion texture requires.
Training was fragile, with the usual GAN sensitivity to hyperparameters and
recurring mode collapse on a few hundred leprosy images. And, most importantly,
the architecture offered no natural mechanism for \emph{spatial} control: the class
label determines what kind of image is produced but not where the lesion sits or
what shape it takes. Mask conditioning, which is straightforward in a diffusion
UNet, was the capability we most needed.

\subsection{Feature extraction and lesion characterization}
\label{sec:features}

A separate line of work aimed at structured lesion description rather than
synthesis. Working from the CO2Wounds-V2 COCO annotations, we built two parallel
labelers. The first was a rule-based computer-vision pipeline that decoded each
polygon to a mask and computed morphological descriptors: lesion count and area,
circularity and eccentricity for shape, solidity for border definition, satellite
detection by connected-component area ratio, and pigmentation class by comparing
mean HSV value inside the lesion against a $10$-pixel dilated ring outside it. The
second used a vision-language model to emit the same schema---shape, border
clarity, border type, color, symmetry, body location---directly from the image.

The intent was a clinically grounded feature vocabulary aligned to the
Ridley--Jopling spectrum, which would eventually support conditional generation by
disease stage. It was set aside for a practical reason: the descriptors that
distinguish leprosy \emph{stages} (hypopigmented macules with sensory loss,
satellite lesions, sloping versus punched-out borders) are largely absent from
CO2Wounds-V2, which depicts chronic ulceration rather than the classifying
morphologies. Building a stage-conditional generator requires a corpus spanning
the spectrum, which does not publicly exist. The vocabulary developed here remains
the natural conditioning interface should such data become available, and we
return to it in Section~\ref{sec:future}.

\section{Evaluation Protocol}

\subsection{Metric}

We evaluate with LPIPS \cite{zhang2018lpips}, which measures perceptual distance
between images as a weighted $L_2$ distance in the feature space of a pretrained
network---here AlexNet, at $256 \times 256$, inputs scaled to $[-1,1]$. LPIPS is a
\emph{distance}: lower means more perceptually similar. It correlates with human
judgments far better than pixel-space metrics such as MSE or PSNR, which is why it
is preferred for generative image evaluation.

For the held-out evaluation we add kernel inception distance (KID)
\cite{binkowski2018kid}, which compares Inception features of two image sets with a
polynomial-kernel maximum mean discrepancy. Unlike FID \cite{heusel2017fid}, KID is
unbiased at small sample sizes, so it is our primary distribution-level number. FID
is reported for completeness but, with $136$ real images, is dominated by
small-sample bias and should not be compared with published values.

\subsection{Kaggle-mask protocol (first run)}
\label{sec:kaggleprotocol}

Generation drew $250$ masks at random (seed $122$) and produced one base-model and
one fine-tuned image per mask under matched seeds. In parallel, $250$ real leprosy
images were sampled and ROI-cropped with the identical procedure. Eight sampled
masks were empty, leaving $242$ images in each set.

Each comparison pairs every anchor image in set $A$ with $5$ randomly chosen
partners from set $B$, giving $242 \times 5 = 1{,}210$ pairs per comparison. It is
important to be precise about what this measures. Because anchors and partners are
always \emph{different} images, each figure is a mean between-image distance: it
characterizes how perceptually spread out two collections are relative to one
another, not how closely any generated image reproduces a specific real one.

\subsection{Anchoring the comparison}
\label{sec:baseline}

Cross-set comparisons such as real versus fine-tuned, fine-tuned versus
fine-tuned, and fine-tuned versus base are dispersion measurements that cannot be
interpreted on their own. A real-versus-fine-tuned distance of $0.716$ (first run)
is neither good nor bad in isolation; it is meaningful only relative to how far
apart two real leprosy images typically are. The protocol therefore includes a
real-versus-real comparison, which supplies that reference scale.

The baseline is computed under the identical protocol as the comparison it
anchors: the same source directory, the same global seed $122$, the same
$250$-image draw, the same ROI margin, the same LPIPS backbone and resolution,
and the same five-partner scheme. This yields the same $242$ anchors and excludes
the same eight images for empty masks (\texttt{IMG671}, \texttt{IMG1297},
\texttt{IMG923}, \texttt{IMG1034}, \texttt{IMG1046}, \texttt{IMG938},
\texttt{IMG896}, \texttt{IMG488}), in the same order, so the baseline and the
cross-set comparison rest on exactly the same anchor set and are directly
comparable.

Because the $1{,}210$ pairs share anchors and are not independent, we report a
$95\%$ confidence interval from a cluster bootstrap that resamples anchor
\emph{images} (5{,}000 replicates) rather than pairs.

This protocol has three confounds: the generation masks come from the
Kaggle \emph{wound} corpus rather than from leprosy lesions; the real reference
images come from the pool used for fine-tuning; and the real reference crops were
taken from the raw photographs, so some still contain the color calibration cards
that were removed from the fine-tuning images. We therefore use it only to compare
the two training runs (Table~\ref{tab:replication}).

\subsection{Held-out protocol}
\label{sec:heldout}

The held-out protocol, applied to the second run, removes those confounds. Each of the $136$ held-out leprosy
lesions supplies the conditioning mask, and its real ROI crop, taken from the
card-removed image as in fine-tuning, forms the real reference set. Each model
generates two images per mask with fixed seeds, shared between models, giving $272$
images per model. A third set runs the base weights with the fine-tuning prompt
(one seed per mask), to separate the effect of the weights from that of the prompt.

\textbf{Set-level LPIPS} uses the five-partner scheme above with the $136$ masks as
clusters ($680$ pairs per comparison), and the same real-versus-real baseline.
\textbf{Paired LPIPS} compares the base and fine-tuned outputs for the same mask and
seed. On its own a paired distance has no scale, so it is set against the distance
between the two seeds of one model for the same mask: if fine-tuning changes an
image less than resampling does, the paired distance is the smaller of the two.
Differences are tested per mask with a Wilcoxon signed-rank test. \textbf{KID and
FID} compare each generated set with the $136$ real held-out crops (Inception
features, KID averaged over $100$ random subsets of $100$ images). All confidence
intervals are $95\%$ cluster bootstraps over masks ($5{,}000$ replicates).

\section{Results}

\subsection{Segmentation}

The Stage 1 network reached a best validation Dice of $0.8758$ and IoU of $0.7993$
at epoch $40$, against training values of $0.9666$ and $0.9381$. The roughly
$0.09$ Dice gap between training and validation indicates mild overfitting, which
is unsurprising given $2{,}208$ images, and validation Dice was still improving
when the epoch budget ran out. These are validation figures on a random split of
the Kaggle training partition; the held-out Kaggle test partition was not scored.
Dice is the soft Dice averaged over validation batches and IoU is computed on
thresholded predictions. The split is by image, and the Medetec portion contains
several photographs per patient, so the validation score is probably optimistic.

\begin{figure}[t]
\centering
\includegraphics[width=\columnwidth]{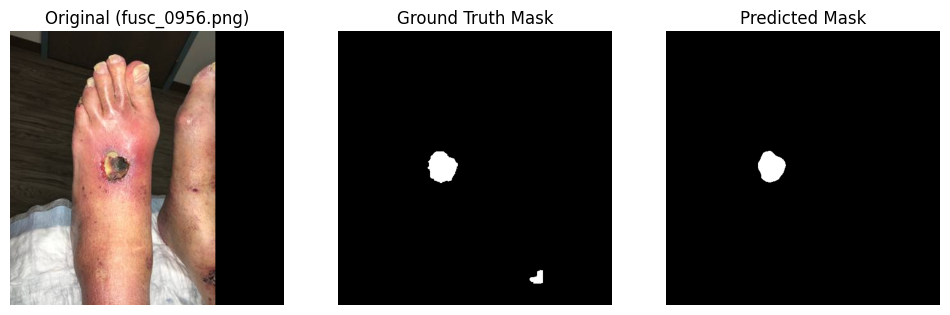}\\[3pt]
\includegraphics[width=\columnwidth]{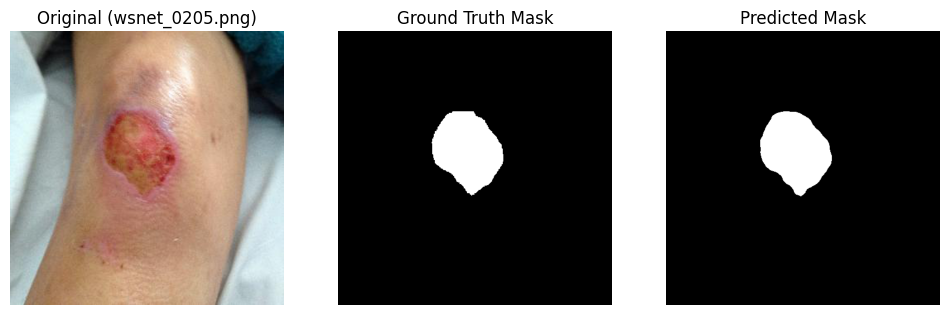}
\caption{Stage 1 segmentation on held-out validation images. Left to right:
input, ground-truth mask, predicted mask.}
\label{fig:seg}
\end{figure}

Qualitatively (Figure~\ref{fig:seg}) predictions track wound boundaries closely on
in-domain images. The practical measure of out-of-domain performance is the
attrition audit of Section~\ref{sec:audit}: $21.9\%$ of AZH masks came back empty,
so segmentation quality degraded substantially on the corpus it was asked to label.

\subsection{Training dynamics}

\begin{figure}[t]
\centering
\includegraphics[width=\columnwidth]{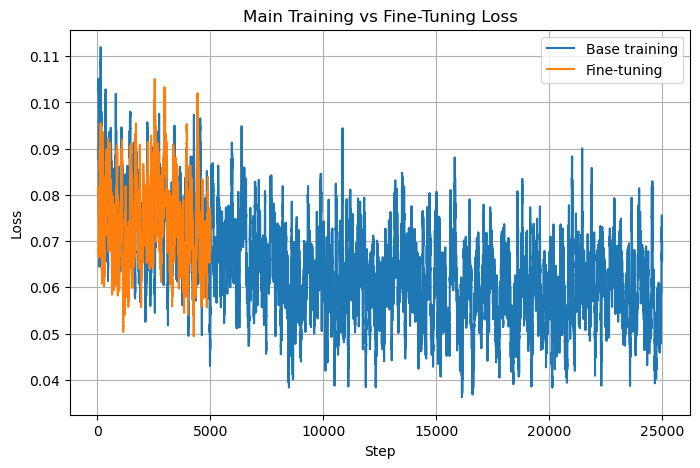}
\caption{Min-SNR-weighted training loss in the first run. Base training (blue)
declines steadily over $25{,}000$ micro-steps. Fine-tuning (orange) remains
approximately flat over $5{,}000$.}
\label{fig:loss}
\end{figure}

Base training reduced the $50$-step smoothed Min-SNR-weighted loss from
approximately $0.080$ to approximately $0.060$ over $25{,}000$ micro-steps, with
the decline steepest in the first $8{,}000$ and continuing gradually thereafter
(Figure~\ref{fig:loss}). Fine-tuning showed a different picture: the smoothed loss
remained near $0.070$ across all $5{,}000$ micro-steps with no clear downward
trend.

We deliberately do not lean on the raw first and last logged values ($0.0408$ to
$0.0168$ for base training, $0.0598$ to $0.0536$ for fine-tuning) because each is
a single stochastic micro-batch at a randomly drawn timestep, and the variance
between adjacent steps exceeds the difference between endpoints. The smoothed
curve is the interpretable signal. The second run behaves the same way: the
mean loss over the first and last $500$ micro-steps falls from $0.0810$ to $0.0606$
in base training and only from $0.0767$ to $0.0714$ in fine-tuning. A flat
fine-tuning loss does not mean the outputs did not change, however.
Section~\ref{sec:lpips} shows that they moved measurably toward the target domain.

\subsection{Qualitative results}
\label{sec:qualitative}

\begin{figure*}[t]
\centering
\includegraphics[width=0.92\textwidth]{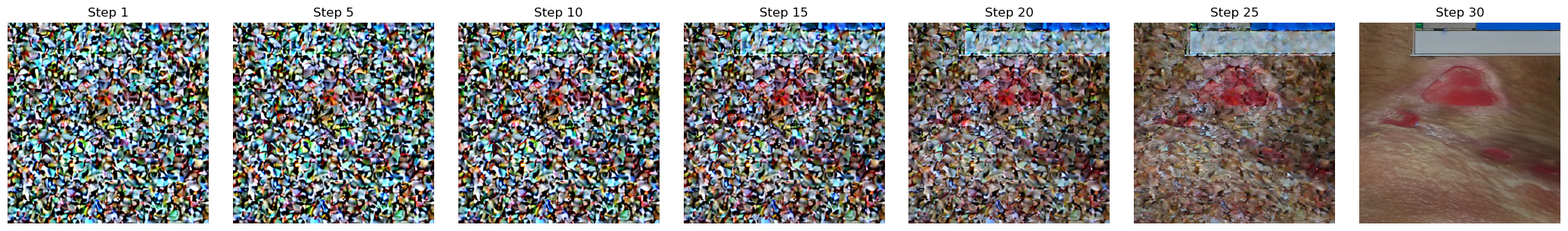}
\caption{Reverse diffusion from Gaussian noise to a synthetic lesion, sampled
every five DPM-Solver++ steps. Global structure resolves early; later steps refine
texture.}
\label{fig:progression}
\end{figure*}

Generated images are convincing at the level of local texture, which is the part of
the problem we consider genuinely solved. Figure~\ref{fig:progression} shows the
reverse diffusion trajectory: coarse color regions resolve into skin structure
within the first third of the schedule, and the remaining steps refine texture.
Outputs reproduce the color vocabulary of real wound photography---red granulation,
yellow-green slough, dark necrosis, peri-wound erythema---along with plausible
surface irregularity and moisture cues, and realistic surrounding skin.

\begin{figure*}[t]
\centering
\includegraphics[width=\textwidth]{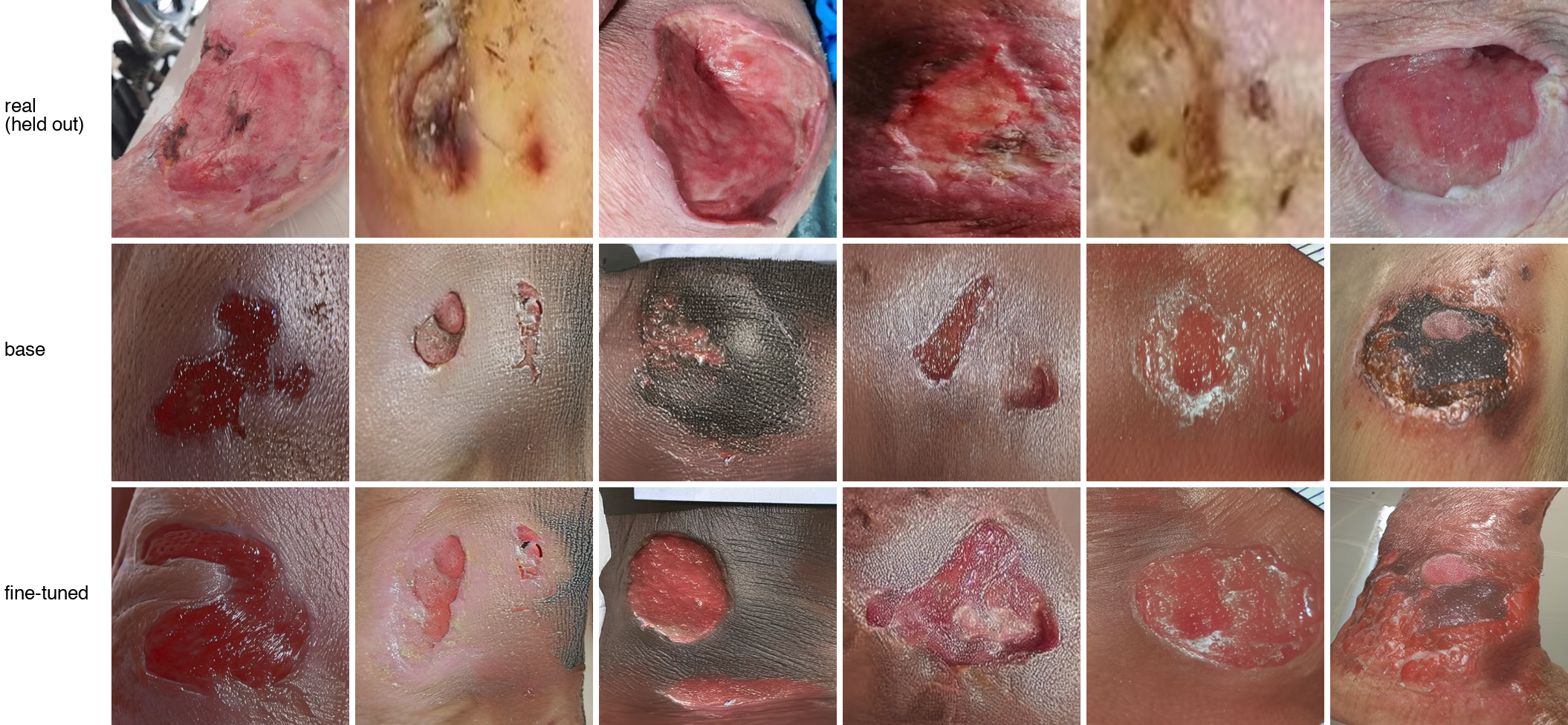}
\caption{Base and fine-tuned models on held-out leprosy lesions. Each
column is one held-out lesion: the real image (top), and the base and fine-tuned
outputs generated from that lesion's mask with the same seed. Fine-tuning keeps the
layout and changes the tissue toward the granulation typical of the leprosy
corpus. These six were selected after visual review of all $272$ held-out outputs;
Figure~\ref{fig:failures} shows typical less successful outputs.}
\label{fig:basevft}
\end{figure*}

Figure~\ref{fig:basevft} shows the base and fine-tuned models on the same
held-out inputs, with the real held-out image for reference. With the input fixed,
the two models usually agree on the overall layout of the image; fine-tuning
changes the tissue itself, toward the bright red granulation and lighter wound
margins that dominate the leprosy corpus. The quantitative counterpart is the
paired LPIPS of Section~\ref{sec:lpips}. Further examples are in
Appendix~\ref{app:examples}.

\begin{figure}[t]
\centering
\includegraphics[width=\columnwidth]{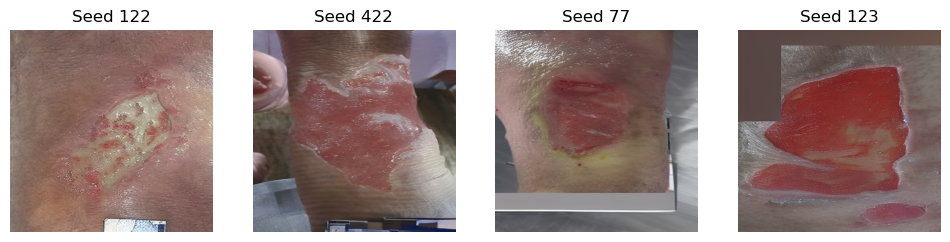}
\caption{Four seeds, one mask, fine-tuned model from the first run. Lesion appearance varies widely,
supporting the diversity result of Section~\ref{sec:lpips}. Shape also varies more
than tight mask adherence would allow, and flat rectangular regions of the kind
audited in Section~\ref{sec:audit} are visible in two panels.}
\label{fig:seeds}
\end{figure}

Sampling one mask under different seeds (Figure~\ref{fig:seeds}) produces
substantially different lesions, which is the qualitative counterpart to the
diversity result below and is evidence against memorization. It also exposes the
main qualitative weakness: lesion shape varies more across seeds than strict mask
adherence would permit, indicating that the conditioning constrains geometry
loosely.

\subsection{Perceptual distance and distribution shift}
\label{sec:lpips}

All numbers in this subsection come from the second run under the held-out
protocol of Section~\ref{sec:heldout}, except Table~\ref{tab:replication}, which
compares the two runs under the Kaggle-mask protocol.

\begin{table}[t]
\centering
\small
\caption{Held-out set-level LPIPS (AlexNet, $256^2$; $680$ pairs from $136$
held-out lesions per comparison). Lower means more perceptually similar. Intervals
are $95\%$ cluster bootstraps over lesions.}
\label{tab:lpips}
\begin{tabular}{lccc}
\toprule
Comparison & Mean & SD & 95\% CI \\
\midrule
real vs.\ real               & 0.667 & 0.084 & [0.658, 0.676] \\
base vs.\ base               & 0.700 & 0.085 & [0.691, 0.710] \\
fine-tuned vs.\ fine-tuned   & 0.734 & 0.056 & [0.727, 0.740] \\
real vs.\ base               & 0.806 & 0.098 & [0.795, 0.818] \\
real vs.\ fine-tuned         & 0.765 & 0.087 & [0.754, 0.777] \\
fine-tuned vs.\ base         & 0.766 & 0.075 & [0.758, 0.773] \\
\bottomrule
\end{tabular}
\end{table}

\begin{table}[t]
\centering
\small
\caption{KID and FID of each generated set against the $136$ real held-out crops.
KID is mean $\pm$ SD over $100$ subsets; lower is closer. FID is biased at this
sample size and is shown for completeness only.}
\label{tab:kid}
\begin{tabular}{lccc}
\toprule
Generated set & $n$ & KID & FID \\
\midrule
base                         & 272 & $0.167 \pm 0.010$ & 285.9 \\
fine-tuned                   & 272 & $0.089 \pm 0.005$ & 203.7 \\
base, leprosy prompt         & 136 & $0.250 \pm 0.008$ & 349.9 \\
\bottomrule
\end{tabular}
\end{table}

\begin{table}[t]
\centering
\small
\caption{Paired LPIPS on held-out lesions: the same lesion mask under both models
or both seeds. Intervals are $95\%$ cluster bootstraps over lesions.}
\label{tab:paired}
\begin{tabular}{lcc}
\toprule
Comparison & Mean & 95\% CI \\
\midrule
base vs.\ fine-tuned, same seed          & 0.544 & [0.529, 0.557] \\
\quad same seed and prompt               & 0.623 & [0.600, 0.644] \\
fine-tuned, seed 1 vs.\ seed 2           & 0.728 & [0.718, 0.739] \\
base, seed 1 vs.\ seed 2                 & 0.685 & [0.669, 0.700] \\
base, own vs.\ leprosy prompt            & 0.350 & [0.328, 0.373] \\
\bottomrule
\end{tabular}
\end{table}

Four findings follow.

\textbf{Fine-tuning moved the outputs toward real leprosy images.} KID against the
real held-out set nearly halves, from $0.167$ for the base model to $0.089$ after
fine-tuning (Table~\ref{tab:kid}). LPIPS agrees: distance from generated to real
images falls from $0.806$ to $0.765$, and the confidence intervals do not overlap
(Table~\ref{tab:lpips}). Measured against the real-versus-real baseline of $0.667$,
the gap between generated and real images shrinks from $0.139$ to $0.098$. The
fine-tuned and base sets are also farther apart ($0.766$) than two samples of the
fine-tuned model are from each other ($0.734$), so the two models produce
distinguishable distributions.

\textbf{The shift comes from the weights, not the prompt.} The base model given the
fine-tuning prompt moves \emph{away} from the real set (KID $0.250$), and with the
prompt held fixed the base and fine-tuned outputs still differ
(Table~\ref{tab:paired}, second row).

\textbf{For a fixed input, fine-tuning changes the image less than a new seed does.}
The base and fine-tuned outputs for the same mask and seed are $0.544$ apart, while
two seeds of the fine-tuned model are $0.728$ apart. Per lesion, the difference is
$-0.185$ (95\% CI $[-0.200, -0.170]$, Wilcoxon $p < 0.0001$); it remains negative
against the base model's seed-to-seed distance ($-0.142$, $[-0.160, -0.123]$) and
with the prompt held fixed ($-0.106$, $[-0.129, -0.083]$; both $p < 0.0001$). The
starting noise largely sets the layout of an image, and fine-tuning changes what is
drawn within it, as Figure~\ref{fig:basevft} shows. Both statements are true at
once: each image changes moderately, and the set as a whole moves measurably toward
the target.

\textbf{The generated set shows no mode collapse.} Fine-tuned images are
$0.734$ apart from one another, against $0.667$ for real held-out images. A
collapsed model would fall well below the real baseline; this one sits above it.

\textbf{Agreement between the two runs.} Table~\ref{tab:replication} applies the
Kaggle-mask protocol to both runs. The real-versus-real baseline, which involves no
model, is identical ($0.672$). The two generated-set distances come out about
$0.06$ higher in the second run: under this protocol the second run's model is more
diverse, and correspondingly farther from the real set. We cannot attribute this
to a single cause; among other differences, the second run fine-tuned on $572$
rather than $708$ pairs. The held-out results above are the ones we rely on.

\begin{table}[t]
\centering
\small
\caption{The Kaggle-mask protocol (Kaggle wound masks; $242$ real reference crops
from raw images) applied to both training runs. Intervals, for the second run, are
cluster bootstraps over anchors.}
\label{tab:replication}
\begin{tabular}{lccc}
\toprule
Comparison & Second run & 95\% CI & First run \\
\midrule
real vs.\ real             & 0.672 & [0.664, 0.680] & 0.672 \\
fine-tuned vs.\ fine-tuned & 0.722 & [0.717, 0.727] & 0.662 \\
real vs.\ fine-tuned       & 0.773 & [0.763, 0.783] & 0.716 \\
\bottomrule
\end{tabular}
\end{table}

\subsection{Less successful outputs}
\label{sec:failures}

\begin{figure}[t]
\centering
\includegraphics[width=\columnwidth]{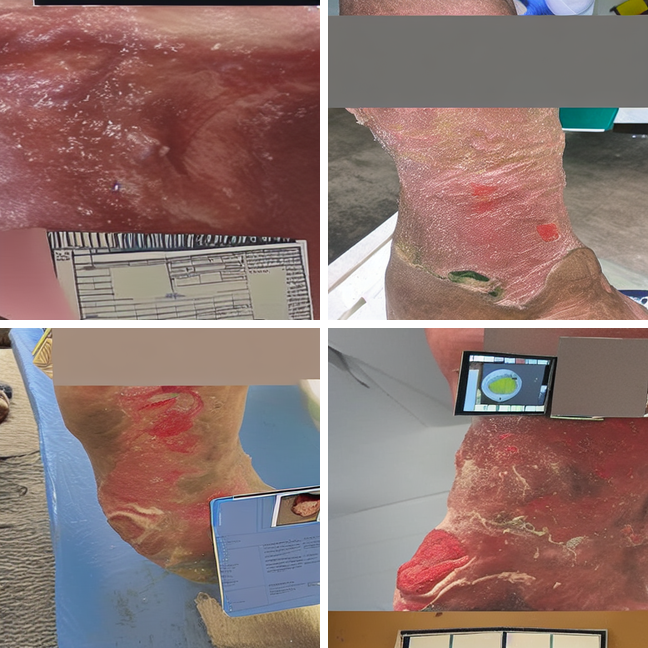}
\caption{Less successful outputs of the fine-tuned model on held-out
lesions: a paper form (top left), a flat gray rectangle (top right), and card- and
screen-like objects (bottom).}
\label{fig:failures}
\end{figure}

Not every output is as clean as those in Figure~\ref{fig:basevft}.
Figure~\ref{fig:failures} shows typical less successful outputs of the fine-tuned
model. Some frame a whole limb in a clinical setting instead of a lesion close-up.
Others contain objects from the photographic setting of the leprosy corpus---paper
forms, screens, card-like objects---or flat gray rectangles of the kind left by
card redaction (Section~\ref{sec:artifacts}). The base model rarely produces either.
Both behaviors are learned from the fine-tuning images, which show such scenes and
contain such rectangles, and they are the most visible remaining weakness of the
model.

\subsection{Artifact propagation}
\label{sec:artifacts}

Section~\ref{sec:audit} established that $57.8\%$ of base training images and
$41.0\%$ of fine-tuning images contain large synthetic flat regions. The natural
question is whether the model learned them.

In the first run, applying a block-level uniformity detector ($16 \times 16$
blocks, per-channel standard deviation below $2$ at $256^2$) to the $18$ generated
images recoverable from our figure archive gives a mean flat-block fraction of
$0.033$, with $5$ of $18$ ($27.8\%$) exceeding a $5\%$ flat-block threshold. The
same detector on the $242$ real ROI crops gives a mean of $0.034$ and $19.0\%$ above
threshold. Eighteen images are too few to establish an elevation over the real-crop
rate. In the second run the question is settled by inspection: flat rectangles
appear in many fine-tuned outputs (Section~\ref{sec:failures}).

This bears on the KID result. Several real held-out crops also contain fill
rectangles, and some still show color cards that the redaction missed. Part of the
fine-tuned model's closer match to the real held-out set may therefore come from
reproducing these regularities of the corpus, not only lesion appearance. What can
be said firmly is that a majority of training images on both sides of the transfer
contain a systematic artifact of preprocessing rather than of anatomy, and that the
model learns it. For anyone assembling clinical corpora from heterogeneous public
sources, this is a failure mode worth checking before training rather than after.

\section{Discussion}
\label{sec:discussion}

\subsection{What transferred, and what did not}

The results separate cleanly into two layers.

Low-level appearance transferred well. Generated images reproduce wound texture,
the clinical color vocabulary, surface irregularity, and peri-lesional skin at a
quality that is convincing on inspection, and the generated set is at least as
diverse as the real leprosy set. Since chronic ulceration in leprosy patients is
visually continuous with chronic ulceration generally, the donor domain was well
chosen: the wound corpora taught the model most of what it needed. Fine-tuning then
did what it was meant to do. With a few hundred images and the down-blocks frozen,
it moved the output distribution measurably toward real leprosy lesions while
keeping the layout set by the starting noise. The flat fine-tuning loss reflects a
loss already near the level base training had reached, not an absence of
adaptation.

Higher-level control did not transfer, and was arguably never established. Lesion
shape varies more across seeds than mask conditioning should permit, and the model
has no handle on severity, disease stage, anatomical site, or lighting. Fine-tuning
also taught it habits of the leprosy corpus that are not lesions: framing a whole
limb, and drawing the objects and flat rectangles present in the training
photographs (Section~\ref{sec:failures}). It renders a plausible lesion; it does not
render a specified lesion in a specified place.

\subsection{Why the conditioning is loose}

Section~\ref{sec:residual} supplies a concrete mechanism. Geometry reaches the
network through a single path: three mask channels concatenated to the latent at
\texttt{conv\_in}, downsampled to $64 \times 64$. That signal must survive the
entire encoder to influence generation, and its influence is mediated by seven
newly initialized input channels competing against four channels carrying the
full pretrained prior. ControlNet \cite{zhang2023controlnet} was introduced
precisely because input-level conditioning underconstrains diffusion models, and
its remedy---injecting conditioning features at every resolution through
zero-initialized convolutions---directly targets this failure. Our multi-scale
encoder was built but, as documented, never engaged. The context hint may add to
this: during training the network sees a blurred copy of the target crop on $70\%$
of steps, a strong layout cue it never receives at sampling time, so it has less
reason to rely on the mask.

The same picture fits the paired result. With down-blocks frozen, adaptation was
confined to the decoder path. Fine-tuning changed the tissue and color drawn within
an image, while the layout---set by the noise and only loosely by the mask---stayed
largely in place.

\subsection{Interpreting the metric honestly}

The LPIPS protocol measures set dispersion, not per-image fidelity. It answers
``how far apart are two collections?'' rather than ``how real does this image
look?''. KID answers a related distribution-level question. Together they support
two claims---no mode collapse, and a measurable shift toward the target after
fine-tuning---but not a claim of photorealism, which needs a clinician preference
study or a reader test.

The held-out protocol removes the main confounds of the original evaluation: the
masks and real references now come from held-out leprosy lesions, and the real
references are card-removed like the fine-tuning images. Two caveats remain. The
split is by image, so held-out lesions can come from patients seen in fine-tuning,
which would flatter the fine-tuned model. And KID may partly reward the fine-tuned
model for reproducing regularities of the corpus, such as fill rectangles, that
the real held-out crops share (Section~\ref{sec:artifacts}).

\subsection{Ethical position}

Synthetic medical imagery invites two reasonable objections, and both have clean
answers here. On privacy: no real pixels enter the generation path, so outputs are
not derived from any individual patient's photograph, though we note that
memorization is never formally excluded without a nearest-neighbor audit, which we
recommend as standard practice. On clinical use: this model is a data-augmentation
and research tool. Its outputs have not been validated by a dermatologist, they do
not span the diagnostic spectrum, and they must not be used for training clinical
judgment or as diagnostic reference material.

\section{Limitations}
\label{sec:limitations}

\textbf{Domain scope.} CO2Wounds-V2 depicts chronic ulceration in leprosy patients,
not the hypopigmented anesthetic macules and plaques on which early leprosy
diagnosis actually turns. The model covers a real presentation but not the
diagnostically decisive one.

\textbf{Evaluation scope.} Two automatic metrics (LPIPS and KID) on $136$ held-out
lesions, with no memorization audit and no expert assessment. FID is unreliable at
this sample size.

\textbf{Patient-level dependence.} $96$ patients supply $764$ images, and the corpus
does not identify them, so the held-out split is by image and the same patient can
appear on both sides. Any stronger fidelity claim requires partitioning by patient.

\textbf{Run-to-run variation.} The held-out results come from a single run. The
difference between the two runs under the Kaggle-mask protocol
(Table~\ref{tab:replication}) shows that run-to-run variation is not negligible.

\textbf{Pseudo-label quality.} $1{,}009$ of $3{,}769$ base training pairs
($26.8\%$) used predicted rather than ground-truth masks, never manually verified.
The $21.9\%$ empty-mask rate on AZH shows these degrade materially under domain
shift.

\textbf{Preprocessing artifacts.} A majority of training images on both sides carry
synthetic flat regions, the color-card redaction is visibly incomplete on some
images, and the fine-tuned model reproduces both.

\textbf{Conditioning mechanism.} The multi-scale residual path was implemented but
not active in the reported runs (Section~\ref{sec:residual}); reported conditioning
is input concatenation only.

\textbf{Segmentation evaluation.} Dice $0.876$ is validation performance on a
random split of the Kaggle training partition, not held-out test performance.

\section{Future Work}
\label{sec:future}

Four steps follow directly, ordered by expected value.

\textbf{Engage the multi-scale conditioning and ablate it.} The residual encoder is
already written. Training two otherwise identical models---with and without the
residual path engaged---would convert Section~\ref{sec:residual} from a caveat into
a controlled experiment and would test the mechanism we propose for loose mask
adherence.

\textbf{Finish the evaluation.} Held-out leprosy masks, paired same-input
distances, and KID are in place. What remains is a patient-level split, which needs
patient identifiers from the dataset authors; a nearest-neighbor memorization
audit; and a reader study. Cleaner card redaction---inpainting instead of a flat
fill---would also remove the artifact the model now learns.

\textbf{Prove downstream utility.} The strongest justification for synthetic data
is that it improves a real task. Training a leprosy wound segmenter on real data
alone versus real plus synthetic, and comparing held-out Dice, would test the
augmentation premise directly. This is the experiment that would most strengthen
the work.

\textbf{Condition on clinical semantics.} The feature vocabulary of
Section~\ref{sec:features}---lesion count, border character, pigmentation, symmetry, site---is the
natural conditioning interface for stage-controlled generation across the
Ridley--Jopling spectrum. Realizing it requires an image corpus spanning that
spectrum, which does not publicly exist and whose construction would be a
contribution in its own right.

\section{Conclusion}

We asked whether a diffusion model pretrained on abundant chronic wound imagery
could be transferred to the severely data-limited domain of leprosy. The answer is
a qualified yes, and the qualification is now quantified.

A mask-conditioned latent diffusion model trained on $3{,}280$ wound crops and
fine-tuned on a few hundred leprosy crops moves measurably toward real leprosy
images: on $136$ held-out lesions, KID falls from $0.167$ to $0.089$ and LPIPS
distance to real images from $0.806$ to $0.765$, with non-overlapping confidence
intervals, and the prompt alone does not explain the gain. The generated set keeps
at least the real set's diversity, so there is no mode collapse. These claims rest
on the real-versus-real baseline; without such a reference, cross-set perceptual distances are unanchored
numbers, and we would most want that point carried forward.

Two weaknesses accompany the result, and both have identified mechanisms. Mask
control is loose, which we trace to geometry reaching the network through input
concatenation alone rather than through the multi-scale path that was implemented
but not engaged. And a majority of training images on both sides of the transfer
carry large synthetic flat regions from letterbox padding and color-card
redaction, which the fine-tuned model learns to draw. Both are actionable.

The broader premise holds up. Chronic wound photography is a viable donor domain
for leprosy-related lesion synthesis, low-level appearance transfers well across it,
and the remaining barrier is semantic control rather than image quality. For
neglected tropical diseases, where data scarcity is the binding constraint and is
unlikely to ease soon, that is a route worth developing further.

\section*{Data and Code Availability}

Code for preprocessing, segmentation, diffusion training, and evaluation is
available at \url{https://github.com/Yusufa09/SyntheticImageGeneration}, including
the training and evaluation code for the held-out results in the \texttt{rerun}
folder. All image
data are from public sources cited above: the Kaggle Wound Segmentation Images
collection \cite{kagglewoundseg}, the AZH Wound Database
\cite{wang2020wound,anisuzzaman2022mobile}, WoundsDB \cite{chronicwounddb}, and
CO2Wounds-V2 \cite{sanchez2024co2wounds}. No new patient data were collected. 

\section*{Ethics Statement}

This study involved only computational analysis of publicly available, de-identified
image datasets. No human participants, vertebrate animals, biological agents, or
hazardous materials were involved, and no institutional review was required. The
synthetic images produced are research artifacts and are not validated for clinical
or diagnostic use.

\section*{Acknowledgments}

Compute was provided by AWS SageMaker and AWS EC2. All experimental results, graphs, figures, and data tables in this
paper were produced by the author from the datasets cited above.

The author used AI assistants in this work. A GPT vision-language model (OpenAI)
located color calibration cards, as described in the Data section. Anthropic's
Claude helped review the code, write the training and evaluation scripts for the
second run, and edit the manuscript. The author checked the code, results,
and text and is responsible for them.

\appendix
\onecolumn
\section{Additional held-out examples}
\label{app:examples}

\begin{figure}[h]
\centering
\includegraphics[width=\textwidth]{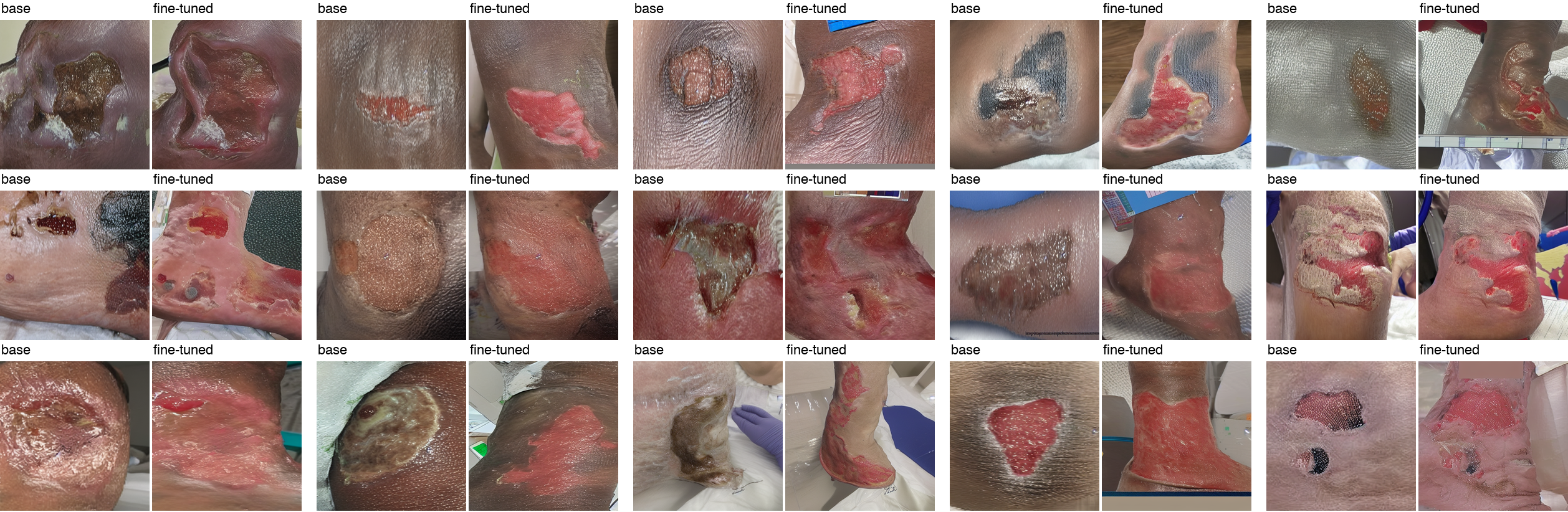}
\caption{Further pairs from the base and fine-tuned models on held-out leprosy lesions. In
each pair the base output (left) and fine-tuned output (right) share the same input
and seed. Selected, like Figure~\ref{fig:basevft}, after visual review of all
held-out outputs.}
\label{fig:more}
\end{figure}

\end{document}